%% file: acl_latex.tex
\pdfoutput=1

\documentclass[11pt]{article}

\usepackage[final]{acl}

\usepackage{times}
\usepackage{latexsym}
\usepackage{tabularx}
\usepackage{float}
\usepackage[T1]{fontenc}

\usepackage[utf8]{inputenc}

\usepackage{microtype}

\usepackage{inconsolata}

\usepackage{graphicx}
\usepackage{booktabs}
\usepackage{xcolor}
\usepackage{colortbl}
\usepackage{highlight}

\title{Exploring the Use of VLMs for Navigation Assistance \\ for People with Blindness and Low Vision}

\author{\textbf{Yu Li}$^\dag$, \textbf{Yuchen Zheng}$^\dag$, \textbf{Giles Hamilton-Fletcher}$^\ddag$, \textbf{Marco Mezzavilla}$^\diamond$,\\ \textbf{Yao Wang}$^\ddag$, \textbf{Sundeep Rangan}$^\ddag$, \textbf{Maurizio Porfiri}$^\ddag$, \textbf{Zhou Yu}$^\dag$, \textbf{John-Ross Rizzo}$^\ddag$ \\
  $^\dag$Columbia University\quad
  $^\ddag$New York University\quad 
  $^\diamond$Politecnico di Milano\\
}

\begin{document}
\maketitle
\begin{abstract}
This paper investigates the potential of vision-language models (VLMs) to assist people with blindness and low vision (pBLV) in navigation tasks. We evaluate state-of-the-art closed-source models, including GPT-4V, GPT-4o, Gemini-1.5-Pro, and Claude-3.5-Sonnet, alongside open-source models, such as Llava-v1.6-mistral and Llava-onevision-qwen, to analyze their capabilities in foundational visual skills: counting ambient obstacles, relative spatial reasoning, and common-sense wayfinding-pertinent scene understanding. We further assess their performance in navigation scenarios, using pBLV-specific prompts designed to simulate real-world assistance tasks. Our findings reveal notable performance disparities between these models: GPT-4o consistently outperforms others across all tasks, particularly in spatial reasoning and scene understanding. In contrast, open-source models struggle with nuanced reasoning and adaptability in complex environments. Common challenges include difficulties in accurately counting objects in cluttered settings, biases in spatial reasoning, and a tendency to prioritize object details over spatial feedback, limiting their usability for pBLV in navigation tasks. Despite these limitations, VLMs show promise for wayfinding assistance when better aligned with human feedback and equipped with improved spatial reasoning. This research provides actionable insights into the strengths and limitations of current VLMs, guiding developers on effectively integrating VLMs into assistive technologies while addressing key limitations for enhanced usability.
\end{abstract}

\section{Introduction}
Navigating the world presents significant challenges for people with blindness and low vision (pBLV). According to the World Health Organization (WHO), approximately 285 million people worldwide are visually impaired, with 39 million living with complete blindness~\cite{world2013media, pascolini2012global}. This number is expected to increase significantly in coming decades. Visual impairment profoundly impacts daily life, especially in the area of mobility. pBLV often encounter difficulties in moving through complex environments safely and efficiently, which limits their independence and ability to participate in myriad activities. While previous assistive technologies for pBLV, such as mobile applications~\cite{10.1007/978-3-030-51517-1_36, nayak2020assistive, gintner2017improving} and wearable devices~\cite{s21041536, kaiser2012wearable, zare2022wearable}, offer essential support, they are often inadequate in providing comprehensive navigation assistance in dynamic and unfamiliar settings. The development of vision-language models (VLMs)~\cite{Du2022ASO, zhang2023visionlanguage, bordes2024introduction} holds significant promise in addressing these limitations and could enhance the quality of life for pBLV.

Recent advancements in large language models (LLMs), such as GPT-3~\cite{NEURIPS2020_1457c0d6} and GPT-4~\cite{openai2024gpt4}, have significantly enhanced the capabilities of natural language processing tasks, including text generation~\cite{10.1145/3649449}, machine translation~\cite{haddow-etal-2022-survey}, and summarization~\cite{10.1162/tacl_a_00632}. Building on the success of LLMs, pre-trained VLMs, such as CLIP~\cite{pmlr-v139-radford21a}, LLaVA~\cite{NEURIPS2023_6dcf277e}, and GPT-4 with vision (GPT-4V)~\cite{openai2024gpt4, 2023GPT4VisionSC, yang2023dawn}, have shown potential in addressing various complex computer vision tasks, from visual question answering~\cite{bioengineering10030380} to image captioning~\cite{9880405}. These models are designed to understand images in context, enabling them to bridge the gap between visual information and natural language description by interpreting scenes holistically, such as recognizing a city park or a birthday party, rather than merely identifying objects.

VLMs have the potential to serve as strong foundations for developing dialogue systems explicitly designed to facilitate navigation tasks for pBLV. These models appear to show strong spatial understanding capabilities, but their actual robustness for high-stakes assistive tasks is largely unverified. Recent studies have investigated the application of VLMs in navigation tasks, highlighting their ability to comprehend scenes, respond to navigation-related questions, and provide detailed descriptions of visual surroundings~\cite{wang2024vlm, hirose24lelan, liu2024objectfinder, azzino20245g}. For example, prior research has shown that VLMs can effectively describe environments to help mobile agents navigate unfamiliar spaces~\cite{liu2024citywalker}. Another line of research has focused on using VLMs to generate grounded instructions for various scenarios, such as searching through kitchens or navigating outdoor settings for pBLV~\cite{merchant2024generating}. However, before these models can be safely deployed, their fundamental capabilities and, critically, their reliability in high-stakes navigation tasks must be rigorously evaluated.

Efforts have been made to evaluate the performance of VLMs. For instance, existing benchmarking datasets like MMMU~\cite{yue2023mmmu, yue2024mmmu} are designed to test models on their knowledge and reasoning abilities. Further, some studies have developed datasets specifically aimed at navigation tasks~\cite{krantz2020beyond, song2024towards, pmlr-v205-li23a, 9577729}, assessing VLMs' capacity to understand distance, directional cues, and environmental features that are crucial for mobility. However, many of these datasets and benchmarking efforts are unrealistic or not explicitly tailored for navigation tasks involving pBLV. Although VLMs have shown considerable capability in everyday visual-language tasks, their overall effectiveness in navigation, such as scene understanding, distance comprehension, and route planning for pBLV, still requires thorough evaluation. This highlights the necessity for a realistic and systematic VLMs assessment to better understand their potential and limitations in assisting with wayfinding for pBLV. In particular, there is a critical need to evaluate the reliability and consistency of these models, as even rare failures can be dangerous in a high-stakes assistive application.

This study provides such a systematic evaluation focused on model reliability. Rather than aiming for a large-scale, `in-the-wild' dataset, where failures can be attributed to many confounding variables, we present a focused, controlled study to benchmark fundamental VLM capabilities and pinpoint specific failure modes. We developed a new, meticulously curated dataset to create controlled, repeatable testbeds. Our methodology is designed to isolate foundational skills (such as object counting, spatial reasoning, and instruction generation) and, most importantly, to rigorously test model consistency by querying each scenario 100 times. This deep, repeated-measures approach allows us to uncover a critical finding: many state-of-the-art VLMs exhibit significant brittleness and inconsistency, where even minor, controlled changes in a scene can lead to drastically different and incorrect outputs. Our work thus provides a solid `first step' and valuable insights into the practical reliability of current VLMs for navigation assistance, identifying key areas for improvement before real-world deployment.

\section{Objectives}
This study aims to evaluate the capabilities of current state-of-the-art VLMs, in performing tasks essential for navigation. Our primary focus is assessing VLMs abilities in fundamental tasks essential for wayfinding: \emph{counting ambient obstacles, relative spatial reasoning,} and \emph{understanding scenes with common-sense wayfinding relevance.}. Through this systematic and controlled evaluation, we seek to highlight both the strengths and critical limitations of VLMs, which are foundational to handling real-world navigation challenges for pBLV. The guiding questions for our research are:
\begin{enumerate}
    \item Do VLMs possess the skills to accurately count objects, understand spatial relationships, and apply common-sense reasoning in navigation tasks?
    \item How reliable and consistent are VLMs when processing controlled variations of the same scene?
    \item What are the strengths and limitations of VLMs in navigation tasks, and how can these insights guide future improvements?
\end{enumerate}
We aim to provide a comprehensive understanding of the current state of VLMs in aiding navigation for pBLV. The findings from this study are expected to offer valuable insights for future research and development, benefiting stakeholders such as VLM developers and providers of assistive-technology services.

\section{Method}
In this study, we explore the role of VLMs in assisting pBLV during navigation, specifically focusing on electronic travel aids (ETAs). ETAs are technologies designed to aid local navigation by detecting obstacles, identifying landmarks, and enhancing spatial awareness. Unlike electronic orientation aids (EOAs), which focus on large-scale route planning (such as GPS-based navigation), ETAs enable users to make real-time decisions and navigate safely within their immediate surroundings.

To assess the feasibility of VLMs for ETA-based navigation, we evaluate their performance in three fundamental navigation skills: counting ambient obstacles, relative spatial reasoning, and common-sense wayfinding-pertinent scene understanding. Counting obstacles is crucial for detecting potential hazards and understanding environmental complexity. Spatial reasoning enables models to interpret object relationships and determine relative positioning, which is essential for guiding users around obstacles and toward safe paths. Common-sense wayfinding-pertinent scene understanding allows models to infer contextual details, such as whether a seat is occupied or if a pathway is accessible, aiding in effective decision-making. By systematically analyzing these core competencies, we aim to determine whether VLMs can reliably support real-time, local navigation assistance for pBLV.

\subsection{Data Collection}
We created a purpose-built dataset\footnote{https://github.com/rizzojr01/vlm\_navigation\_eval.git} to test various VLMs under controlled conditions. The full dataset contains 150 unique images featuring diverse indoor and outdoor scenes with chairs, benches, and sofas serving as the primary objects. These scenes include open outdoor spaces as well as complex indoor environments. To ensure stability during the experiments and maintain control over the variables, we used a fixed-camera position while systematically modifying specific attributes within the scenes. These adjustments included variations in the number and style of chairs, the placement of obstacles along the path at various stances, and subtle background details. By isolating and precisely adjusting these variables while keeping other elements constant, we achieved a more comprehensive and accurate evaluation of the models' capabilities.

We have publicly released this dataset to advance research in navigation technologies, particularly for individuals with pBLV. This dataset will provide essential support for evaluating and improving VLMs' spatial understanding and path-planning capabilities. It not only lays a solid foundation for subsequent studies but also paves the way for practical applications in assistive technologies, contributing to the development of more robust and efficient navigation solutions. For the experiments in this paper, we selected a focused subset of these scenes, which are detailed in the following sections. To rigorously evaluate model reliability and consistency, each of these unique scene-prompt combinations was queried 100 times, for a total of 3,400 evaluations.

\subsection{Fundamental Capabilities}
To evaluate the core capabilities of VLMs in navigation tasks, we focused on three sub-tasks: counting ambient obstacles, relative spatial reasoning, and common-sense wayfinding-pertinent scene understanding. These sub-tasks are integral to the overall effectiveness of a navigation system designed for pBLV. By systematically assessing the performance of VLMs in these areas, we aim to identify their strengths and weaknesses in providing reliable navigation assistance. The following sections offer detailed discussions on the evaluation methods and findings for each sub-task.

\subsubsection{Counting Ambient Obstacles}
We collected a series of images from scenes featuring varying numbers of chairs, ranging from 1 to 6, to assess the counting abilities of VLMs. Each scene had different layouts and chair counts to represent a spectrum of situations, from simple to complex. We asked various VLMs to count the number of chairs in each scene. We recorded their accuracy to determine if their outputs accurately reflected the actual number of chairs present in the images. This testing process allowed us to systematically assess the ability of VLMs to perceive and quantify environmental elements, a critical skill for ETAs that must detect obstacles and interpret scene complexity to support real-time navigation for pBLV. This sub-task used a set of 8 unique scenes.

\subsubsection{Relative Spatial Reasoning}
Our evaluation of relative spatial reasoning examines how variations in object positioning and appearance impact the models’ ability to provide accurate spatial feedback. The test consists of images featuring two chairs placed at different distances from the viewpoint, with some scenarios presenting chairs of the same color and others incorporating different colors. The models are prompted with the question: ``Which chair is closer to the viewpoint?'' This setup evaluates whether VLMs can assess spatial relationships independently of object appearance or color differences. For ETAs, precise spatial reasoning is crucial to ensure that navigation guidance is based on object positions rather than visual attributes. By incorporating a range of test cases, we analyze the models’ ability to consistently and accurately interpret spatial relationships, providing insights into their reliability for real-world navigation support. This sub-task used a set of 8 unique scenes.

\subsubsection{Common-sense Wayfinding-Pertinent Scene Understanding}
Common-sense reasoning enables systems to interpret contextual information in ways consistent with human understanding, which is crucial for providing accurate and meaningful guidance across myriad scenes. For example, determining whether a seat is vacant requires recognizing the presence of a chair and identifying contextual cues indicating occupancy, such as items like a coat placed on the chair. To evaluate the common-sense reasoning capabilities of VLMs, we designed tests involving various scenarios where seat occupancy was implicitly represented by personal items. The models were prompted with the question: ``Are there any vacant seats in this image?'' To clarify the definition of a vacant seat, we provided the following hint: ``A ‘vacant seat’ refers to a seating option (e.g., chair, bench, or couch) that is unoccupied by any person and does not have any personal items placed on it or on the corresponding desk, table, or surface. Answer yes or no before providing details.'' This setup tested the models’ ability to detect objects on chairs and infer seat occupancy across diverse scenes  based on contextual clues. By systematically varying the scenarios and analyzing the responses, we assessed the strengths and limitations of the models’ common-sense reasoning capabilities, offering insights into their practical utility for scene understanding. This sub-task used a set of 5 unique scenes.

\subsection{Evaluating VLMs for the Navigation Task}
\label{method_navigation}
To assess the effectiveness of VLMs in navigation tasks, we conducted a structured evaluation focusing on their ability to support ETAs. Our evaluation examines how well these models assist in identifying destinations, planning routes, and recognizing obstacles, which are key components of real-world navigation assistance for pBLV. The following steps outline our approach to systematically measuring model performance in these tasks.

\paragraph{Scenario Selection} To evaluate the models’ performance, we designed a diverse set of navigation scenarios with varying levels of complexity, categorized based on the number of obstacles, spatial arrangement of target objects, and the level of ambiguity in the environment. Simple scenarios featured clear, unobstructed paths with minimal distractions, while moderate scenarios introduced partial obstructions or closely placed objects requiring spatial differentiation. Challenging scenarios involved cluttered environments, multiple potential target objects, or obstacles requiring precise route planning. To systematically analyze factors influencing navigation accuracy, we maintained a consistent baseline scenario and introduced controlled variations in obstacles, distances, and target objects. This approach allowed us to assess the adaptability and effectiveness of VLMs under different conditions, using a standardized system prompt and navigation queries with increasing complexity.

\paragraph{Metrics} Responses were not given a single ``correct'' score. Instead, they were independently rated on three criteria:
\begin{itemize}
    \item Destination: Did the model accurately identify and guide the user to  the correct destination?
    \item Route: Did the model correctly provide an optimal route?
    \item Obstacles: Did the model correctly detect obstacles and warn the user to watch out for them?
\end{itemize}

\paragraph{Human Evaluation} The reliability of a model’s navigation ability was evaluated by analyzing the VLMs’ outputs on each metric. To ensure a comprehensive assessment, we asked two graduate students from Columbia University to act as human annotators; both students had no direct involvement in the project so they could label the outcomes in an unbiased manger strictly based on defined metrics. Annotators were provided with detailed instructions and labeling guidelines to ensure consistency and accuracy across all results.

\section{Results}
In this section, we present the results of our evaluation of the navigation capabilities of various VLMs for pBLV, including the fundamental capabilities (counting ambient obstacles, relative spatial reasoning, and common-sense wayfinding-pertinent scene understanding) and the holistic navigation tasks. To gain a broader perspective, we utilize a range of closed-source models (GPT-4V, GPT-4o, Gemini-1.5 Pro, and Claude-3.5-sonnet) as well as open-source models (Llava-v1.6-mistral-7B and Llava-onevision-qwen2-7B) for comparison. A comparative analysis of these models' performances on various fundamental tasks can offers unique insights into their strengths and limitations, helping to determine their suitability for wayfinding support.

\subsection{Fundamental Capabilities}
Here, we present the results of evaluating three fundamental capabilities of various VLMs. To ensure randomness and assess the default capabilities of each model in original form, we utilized default decoding strategies. Each model was tested 100 times per image, and we calculated the accuracies of the output results.

\subsubsection{Counting Ambient Obstacles}
\input{figures/counting}
\input{tables/counting}
We evaluate the counting capabilities of various VLMs using test images containing from one to six chairs. Accuracy is the proportion of the claimed number of chairs in the outputs that matched the actual ground-truth values. We also record the mean and variance of predictions to assess the stability of each model's performance. The test results are summarized in Table~\ref{tab:counting}. 

Among all the models, GPT-4o demonstrated the highest stability and accuracy. In low-density scenes with one, two, or three chairs, it achieved $100\%$ accuracy and maintained strong performance even in more complex scenarios. Across all cases, from one chair to six chairs, its accuracy exceeded $77\%$, showcasing excellent robustness in handling simple and complex spatial distributions.

Other models exhibited notable limitations, particularly in high-density or unevenly arranged scenes. While Gemini-1.5 Pro and GPT-4V perform well in low-density scenarios, their performance deteriorates significantly as scene complexity increases. For instance, Gemini-1.5 Pro achieves only $7\%$ and $14\%$ accuracy in two cases with three chairs and just $25\%$ accuracy in the case with six chairs. The high variance in its predictions further indicates instability. Similarly, GPT-4V struggled in complex scenes, achieving $3\%$ accuracy in the scene with four chairs and only $29\%$ in the case with six chairs. Open-source models such as Llava-qwen and Llava-mistral regularly fail even in the simplest one-chair case, with accuracies of just $38\%$ and $58\%$, respectively.

Claude-3.5-sonnet performs flawlessly in scenes with uniformly distributed chairs, such as four or six chairs, achieving $100\%$ accuracy with zero variance. However, its accuracy drops to $0\%$ in scenarios where chairs are unevenly distributed, such as three or five chairs, with a chair positioned directly in front or to the left. Notably, these extreme results (perfect performance or complete failure) were not influenced by the model's temperature parameter settings in our experiments. We used the default temperature of the Claude API, which was set to 1. The temperature parameter in the API can be adjusted from 0 to infinitely large values, where 0 makes the model behave deterministically, and higher temperatures introduce more variability in the responses. This observation suggests that the results are due to Claude-3.5-sonnet’s inherent spatial reasoning ability rather than a parameter-related bias, highlighting specific blind spots that could hinder its applicability in navigation tasks.

\subsubsection{Relative Spatial Reasoning}
\input{figures/spatial_reasoning_relative}
\input{tables/comparative}

\input{figures/spatial_reasoning_analysis.tex}
To evaluate the spatial reasoning abilities of various VLMs, we conducted relative distance estimation tests using images containing two objects. The models were tasked to determine which object was closer. These images include objects of different colors and styles, with positions systematically varied. Each test was repeated 100 times, and the results are presented in Table~\ref{tab:comparative}.

Among all the models, GPT-4o consistently achieved the highest accuracy across all test scenarios, with scores close to or at 100\% in most cases. This demonstrates its exceptional ability to adapt to different spatial configurations and lighting conditions. For instance, GPT-4o maintained perfect accuracy (100\%) in ``Case 1,'' ``Case 3,'' ``Case 4,'' and ``Case 6,'' and nearly perfect (99\%) in the flipped version of ``Case 2.'' These results highlight its robustness for tasks requiring precise spatial reasoning.

Other models exhibited biases in their results based on the relative positions of the chairs. Specifically, a strong positional bias was observed in certain models. For example, GPT-4V, Llava-qwen, and Claude-3.5-Sonnet achieved significantly higher accuracy when the chairs were of the same style and the closer chair was on the left oddly. For cases where the left chair was closer (Case 1 and Case 2 flipped), these models achieved accuracies of 97\% and 81\% for GPT-4V, 61\% and 74\% for Llava-Qwen, and 77\% and 30\% for Claude-3.5-Sonnet. Conversely, their performance declined when the closer chair was on the right, as seen in Case 1 flipped and Case 2, with accuracies of 61\% and 13\% for GPT-4V, 50\% and 60\% for Llava-Qwen, and 12\% for both cases for Claude-3.5-Sonnet. On the other hand, Gemini-1.5 Pro showed a bias towards perceiving the right chair as being closer. It achieved significantly higher accuracy in scenarios where the right chair was indeed closer, with accuracy rates of 72\% for Case 1 flipped and 53\% for Case 2. In contrast, when the left chair was closer, the model's accuracy dropped to 36\% for Case 1 and 47\% for Case 2 flipped.

To better understand the discrepancies in the outputs of these models, we analyzed the distribution of keywords used in their responses, as illustrated in Figure~\ref{fig_spatial_analysis}. We first evaluated whether the models provided correct spatial reasoning answers and then, within the correct responses, examined whether the answers were optimal for pBLV applications. Optimal answers use spatial terms (e.g., ``left'' and ``right''), which are crucial for effective navigation assistance. Suboptimal answers, while still correct in reasoning, rely on color-based descriptors (e.g., ``orange'' and ``yellow''), which are less useful for pBLV. GPT-4o demonstrates superior performance, consistently providing correct responses and frequently employing spatial terms to describe relative positions. However, it tends to rely on color-based words in its answers when the objects are of different color.

In contrast, other models, such as GPT-4V and Claude-3.5-Sonnet, exhibit a bias towards answering ``left'' over ``right,'' with significant performance gaps between cases involving left and right positioning. Similarly, Gemini-1.5 Pro shows a bias towards answering ``right” over ``left.'' Regarding accuracy, all models except GPT-4o struggle to consistently provide correct spatial reasoning answers, with a significant portion of their responses relying on color-based cues. These findings underscore the need to reduce spatial biases in existing VLMs and align them more closely with pBLV-specific requirements, emphasizing spatial descriptors over less practical visual attributes.

In summary, GPT-4o proved to be the most reliable model for relative spatial reasoning tasks, demonstrating high accuracy and consistency across diverse scenarios. In contrast, other models underperformed in specific conditions and exhibited significant instability in complex setups. We attribute this to biases in their training data and poor alignment with task-specific instructions, making them less suitable for navigation tasks designed for pBLV.

\subsubsection{Common-sense Wayfinding-Pertinent Scene Understanding}
\label{res_commonsense}
\input{figures/commonsense_reasoning}
\input{tables/commonmense}

To assess the common-sense reasoning abilities of VLMs, we developed a series of test cases that featured scenes with varying levels of complexity. We started with a vacant chair as a baseline and then introduced various objects, such as coats, backpacks, and laptops, arranged in different configurations to simulate occupied chairs. Each model was tested 100 times to determine whether a seat was unoccupied or occupied, and the results of their accuracy are summarized in Table~\ref{tab:commonsense}.

GPT-4o demonstrated the highest overall performance, achieving $100\%$ accuracy in most scenarios. This included correctly identifying a vacant chair and cases where a coat was placed on or hung from the chair. Even in more complex configurations, such as the ``Laptop-Backpack'' scenario, GPT-4o maintained a $95\%$ accuracy rate, showcasing its strong ability to interpret implicit cues and reason about object occupancy. However, it faces challenges in the ``Laptop-Backpack-Alt'' case ($7\%$ accuracy), where the backpack hangs off of the chair and the laptop is arranged differently. This highlights its limitation in handling perspective variations within complex setups.

Gemini-1.5 Pro also performed well in more straightforward scenarios, achieving $100\%$ accuracy in the coat tasks. However, its performance dropped significantly in more complex scenarios, such as the ``Laptop-Backpack'' and ``Laptop-Backpack-Alt'' cases, where it achieved only $42\%$ accuracy. Notably, Gemini was the only model that failed to identify the vacant chair case perfectly ($72\%$ accuracy), highlighting a gap in its basic understanding of unoccupied seats. In contrast, GPT-4V, the Llava series models, and Claude-3.5-Sonnet consistently performed poorly across all tested cases involving occupied chairs. Their accuracy was particularly low in tasks requiring implicit state recognition or reasoning about complex object arrangements. While Llava and Claude-3.5-Sonnet succeeded in identifying a vacant chair, they failed in nearly all other cases, underscoring their limitations in handling complex common-sense reasoning.

In summary, GPT-4o is the best model for common-sense reasoning tasks, although it still needs improvement for complex common-sense reasoning. Gemini-1.5 Pro showed strong performance in more straightforward tasks but had moderate accuracy in more complex cases. By contrast, GPT-4V, the Llava series, and Claude-3.5-Sonnet faced significant challenges with common-sense reasoning, highlighting major limitations in their ability to understand implicit and contextual cues.

\subsection{Navigation Task Testing}
We evaluated the model's performance using a specifically designed system prompt and assessed its ability to navigate to a vacant seat under various user queries. This section includes the results of our systematic testing, and an analysis of the model's outputs based on an annotation scheme we developed to categorize potential errors.

\subsubsection{Navigation Task Testing Results}
\label{res_navigation}
\input{figures/navigation_all}
\input{tables/navigation}

We evaluated the performance of VLMs on navigation tasks by assessing their ability to iteratively guide a user toward a vacant chair, providing step-by-step instructions under various conditions. Five models were tested across 13 cases, including classroom and office settings. The evaluation focused on three criteria: the ability to identify the destination, determine the route, and recognize obstacles. Each test case included three user queries, as outlined in Section~\ref{method_navigation}, with each query executed 10 times, resulting in 390 outputs per model. To conduct the human evaluation, we had two graduate student annotators from Columbia University to label the outputs based on the three metrics defined in Section~\ref{method_navigation}. The annotators received training through a pilot study and achieved a Cohen’s kappa value of 0.83 across all annotations, indicating substantial agreement.

As shown in Table~\ref{tab:navigation}, the models demonstrate varying levels of effectiveness across the three metrics. GPT-4o and Claude-3.5-Sonnet exhibit robust accuracy, with both models scoring over 80\% in all metrics. GPT-4o particularly excels in route accuracy (92\%) and obstacle recognition (84\%), while Claude-3.5-Sonnet performs better in destination identification (85\%). Llava-Mistral outperforms Llava-Qwen across all three metrics, but neither model performs well on the navigation task. By contrast, Gemini-1.5 Pro scores only 13\% in route planning, indicating a significant limitation in its ability to plan routes for navigation tasks effectively.

In summary, GPT-4o and Claude-3.5-Sonnet are the most reliable models for high-stakes navigation tasks, providing consistent and accurate guidance even in complex scenarios. The Llava models, while less capable overall, may be better suited to basic navigation tasks with lower accuracy requirements. Gemini-1.5 Pro, however, appears unsuitable for navigation tasks due to its poor performance. In our supplementary materials, we explore additional factors that may have affected VLM performance, including the spatial resolution of each input image used by each VLM (Appendix~\ref{app:resolution}), which could obscure critical visual details at both low and high resolutions, and the influence of system-prompt design on how models interpret user queries and objectives (Appendix~\ref{app:system_prompt}). Additionally, we provide a detailed case study in Appendix~\ref{app:case_study} to illustrate how VLMs handle navigation tasks. These analyses provide context for the observed results and offer insights into potential areas for further optimization.

\input{tables/discussion}

\section{Discussion}
\subsection{Summary of Findings}
This study provides an in-depth evaluation of VLMs for navigation tasks tailored to pBLV, assessing their fundamental skills and performance in practical navigation scenarios. As shown in Table~\ref{tab:discussion}, our results highlight notable differences in model capabilities, with GPT-4o consistently outperforming other models in counting, spatial reasoning, and common-sense scene understanding. It demonstrates substantial accuracy and stability, particularly in high-density environments and complex spatial configurations. In contrast, GPT-4V, Gemini-1.5 Pro, and the Llava series exhibit significant limitations, particularly in handling uneven object distributions or implicit reasoning, revealing gaps in their spatial reasoning and adaptability. In navigation tasks, GPT-4o and Claude-3.5-Sonnet emerge as the most reliable models, excelling in destination identification, route planning, and obstacle recognition. Their performance suggests suitability for high-stakes assistive applications. Conversely, Gemini-1.5 Pro and the Llava models struggle with route planning and obstacle detection, highlighting key weaknesses in their navigation capabilities. While GPT-4o shows strong potential for real-world deployment, further refinements are necessary, particularly in aligning responses with human expectations and reducing inconsistencies across diverse environments.

Our findings reinforce the broader challenges in adapting VLMs for pBLV navigation assistance, particularly biases in spatial reasoning, inconsistencies in object identification, and limitations in scene understanding. While state-of-the-art models show promise, further advancements in model robustness, image resolution processing, and system-prompt optimization are essential to enhance their reliability in real-world navigation applications.

\subsection{Applying VLMs to Realistic Scenarios}
A primary limitation of this study is the controlled, static nature of our dataset. This was a deliberate methodological choice, as our goal was to isolate fundamental VLM failure modes and rigorously test for consistency. However, this focused scope means our findings do not fully represent the complexity of real-world navigation, which involves dynamic environments, variable lighting, motion blur, and unpredictable obstacles. Therefore, the performance and, more importantly, the model inconsistencies we observed in our controlled settings may not directly generalize. Significant future work is needed to bridge this gap and test whether these fundamental ``brittleness'' issues are amplified in more dynamic scenarios.

Adapting VLMs for real-world navigation scenarios also presents challenges beyond our experimental settings. Our findings highlight that even top-performing models are not fully reliable. For instance, while GPT-4o performed best overall in our evaluation, it was not immune to failures on tasks requiring precise spatial reasoning or implicit understanding. Our key finding is this `brittleness', that even the most advanced models can fail in seemingly simple, controlled scenarios. This is a critical issue for a high-stakes assistive application, as even a low failure rate is unacceptable. These difficulties highlight significant architectural and alignment issues that limit the broader applicability of these models.

One important consideration for real-time assistive systems is response time. In practical applications, even small delays can significantly impact usability, particularly in time-sensitive navigation scenarios. While open-source models deployed on dedicated servers can be optimized for lower latency, closed-source models like GPT-4 and Claude often rely on external APIs. This reliance introduces unpredictability, as response times can vary based on factors such as server load, network latency, and access rate limitations. Such variability can present challenges in ensuring a seamless user experience, underscoring the importance of minimizing delays and ensuring consistent performance for real-world deployment of assistive technologies.

A critical challenge for current VLMs lies in their ability to maintain spatial coherence when processing input images of varying resolutions. Techniques like AnyRes~\cite{liu2024llavanext} enable models to handle different image sizes, but they can inadvertently disrupt spatial relationships between objects, affecting tasks that rely on precise spatial reasoning. Maintaining spatial integrity is essential for complex navigation tasks, highlighting the need for improved resolution-handling techniques.

Another issue is the mismatch between model outputs and human feedback for navigation-specific tasks. While GPT-4o demonstrates improved alignment and performs well in interpreting and responding to user instructions, other models, such as Gemini-1.5 Pro and the Llava series, often struggle to meet human expectations in our navigation task evaluation. This discrepancy indicates that current VLMs may not be adequately trained for tasks that require precise image understanding and interpretation based on user input. Techniques like reinforcement learning from human feedback~\cite{ouyang2022training, wang2024} or direct preference optimization transferred~\cite{rafailov2024direct, wang-etal-2024-mdpo} from LLMs have been shown to enhance model performance and adaptability in various applications by aligning image understanding tasks with human feedback.

While models like GPT-4o showcase preliminary promise, our study underscores that adapting VLMs to realistic scenarios requires addressing these critical limitations. By refining architectural approaches to preserve spatial relationships and aligning models more effectively with human expectations, these systems may begin to lay the groundwork for practical utility. However, our findings confirm that significant challenges in fundamental reliability and alignment remain before they can be considered safe or effective for real-world deployment.

\subsection{Application to Assistive Technology}
The application of VLMs in assistive technologies for pBLV has grown significantly, with numerous works exploring their potential to enhance navigation, object detection, and scene understanding. Systems such as \textit{Be My Eyes}\footnote{https://www.bemyeyes.com} have incorporated VLMs to interpret still visual scenes from user-provided images, aiding users in tasks like object identification and obstacle avoidance. Foundational models like GPT-4V have been applied to vision-language planning tasks~\cite{hu2023look, wake2024gpt}, demonstrating potential for assistive applications such as route planning and spatial reasoning. Despite these advances, aligning model predictions with user intent remains a key challenge.

Established wearable assistive technologies have incorporated various AI-driven functionalities to support pBLV. Devices like the Envision Glasses\footnote{https://www.letsenvision.com/glasses} utilize AI to provide functionalities such as text recognition, object identification, and scene description, offering users auditory feedback to interpret their surroundings. Similarly, the OrCam MyEye~\cite{amore2022efficacy}, which has been commercially available for many years, is a compact device that attaches to eyeglasses, employing a combination of specialized AI-driven algorithms for tasks like text reading, face recognition, and product identification. While many of these features rely on task-specific AI technologies, scene description capabilities often draw on VLMs to generate richer contextual information, such as spatial relationships and environmental details, which complement other assistive functionalities. Research initiatives like 5G Edge Vision~\cite{Azzino_2024} are further exploring the integration of VLMs with wearable devices to provide real-time visual scene processing and navigation assistance. By leveraging 5G connectivity, these systems aim to offload the intensive computational required for VLMs, enabling more efficient and responsive support for pBLV. This integration highlights the growing role of VLMs in supplementing existing assistive technologies with enhanced environmental understanding.

Collectively, these efforts underscore the transformative potential of VLMs in assistive technologies. Future directions should improve model alignment with user-specific goals, integrate multimodal features such as tactile and auditory feedback, and enhance real-time adaptability to diverse environments. By addressing these challenges, VLMs can unlock new opportunities to empower pBLV to navigate and interact with their surroundings.

\section{Conclusion}
In summary, advanced VLMs have demonstrated preliminary promise in supporting navigation tasks for pBLV, but our systematic evaluation reveals they still exhibit critical limitations, inconsistencies, and biases that restrict their utility in real-world scenarios. Our study highlights key discrepancies in model performance. While GPT-4o emerged as a relatively strong performer, our findings also show that even this state-of-the-art model is not immune to `brittleness,' failing in controlled tasks that are foundational to navigation. Other models, such as Gemini-1.5 Pro and the Llava series, struggle more significantly in these areas. These findings provide actionable insights for assistive technology developers, highlighting that current models are not yet reliable for deployment in high-stakes applications. They also provide clear direction for VLM developers seeking to refine their models. Future advancements must focus on addressing the specific weaknesses identified in our study, such as improving fundamental spatial reasoning reliability, reducing output biases, and enhancing alignment with human feedback. By addressing these foundational issues, researchers can move closer to creating truly reliable and effective navigation assistance systems tailored to the needs of pBLV.


\bibliography{custom}
\appendix

\section{Appendix}
\label{sec:appendix}
\subsection{Impact of Resolution} 
\label{app:resolution}
\input{tables/resolution-1}
\input{tables/resolution-2}
\input{tables/resolution-3}

Image resolution plays a crucial role in VLMs for navigation tasks. Many state-of-the-art VLMs are pre-trained on datasets with fixed image sizes and utilize methods like AnyRes to handle different resolutions during inference. To evaluate the impact of resolution, we tested different input image resolutions across counting, spatial reasoning, and common-sense reasoning tasks. By comparing the results from original and adjusted resolutions, we observed that resolution adjustments can significantly optimize the accuracy and stability of specific models. Therefore, we selected the optimal resolution for each model in our experiments.

The results for the VLMs at different resolutions are presented in Tables \ref{tab:resolution-1}, \ref{tab:resolution-2}, and \ref{tab:resolution-3}. In spatial reasoning tasks, reducing the resolution to 288×384 significantly improves the accuracy of Llava-Qwen. In contrast, changes in resolution have little effect on the performance of all models for common-sense reasoning tasks, as neither resolution succeeds in completing the task.

Based on the experimental results, Llava-mistral shows the most stable performance at its original resolution, achieving high accuracy in counting and spatial reasoning tasks. Llava-qwen performed best when adjusted to a resolution of 288×384, particularly excelling in complex multi-object scenes. Claude-3.5-sonnet does not show significant improvement with resolution adjustments, as its original resolution is already close to optimal for most tasks. Therefore, for navigation tasks, we used the original resolution for both Llava-mistral and Claude-3.5-sonnet, while adopting the 288×384 resolution for Llava-qwen.

\subsection{System Prompt Design}
\label{app:system_prompt}
In our evaluation of VLMs' navigation capabilities, we focused on its ability to navigate to a vacant chair. Recognizing the significant influence that well constructed prompts can have on a language model's performance, we created a specific prompt system for this task. Our goal was to address all essential aspects to improve VLMs' precision and efficiency in navigation. The detailed prompt system used in our assessments is shown in Figure~\ref{fig_system_prompt}. On the right side of the figure, we list the issues that each part of the prompt is designed to address. By incorporating these specific prompts, we found that VLMs effectively mitigates common problems in navigation tasks. Below we will use GPT-4V to explain the significant impact of prompts on language model performance

Based on the results of our system testing, we can make several important observations about the performance of VLMs in navigation tasks. In our preliminary tests, we found that various VLMs have strong capabilities in comparing the distances of objects of different styles, showing a good understanding of relative spatial relationships. However, it lacks certain capabilities that are critical for practical navigation, such as accurate counting and pixel-level object localization. For example, while GPT-4V can accurately identify the relative distances between objects, it has difficulty accurately counting as the number of objects increases, and cannot accurately localize objects in the image. In addition, the model exhibits biases, such as consistently leaning to the left when determining which object is closer, and misinterpreting common sense scenarios, such as a backpack on a seat indicating that the seat is occupied.

Our navigation task tests show that VLMs currently have difficulty handling navigation tasks. This is due to its lack of basic capabilities. For example, our experiments clearly show that GPT-4V has difficulty estimating distances and generally needs to make common sense reasoning more accurately. These limitations are consistent with our conclusions from the basic ability tests, thus strengthening the validity of our results. Unfortunately, these shortcomings prevent GPT-4V from becoming a reliable component for building dialogue systems for pBLV navigation tasks.
\input{figures/system_prompt}

The input to GPT-4V consists of two main parts: the system prompt and the user query. The system prompt informs the model how to shape its output, including style, length, and level of detail. The user query instructs the model to execute the instruction, such as understanding the input images. In our evaluation, the design of the system prompt is crucial in guiding GPT-4V's responses to provide effective navigation assistance. A well-crafted system prompt significantly impacts the model's ability to offer accurate, relevant, and contextually appropriate guidance. Therefore, we included several key aspects in our system prompt to optimize GPT-4V for our navigation task for pBLV.

\paragraph{Incorporating Navigation Starting Point}
Our system prompt design includes the starting point for navigation, which informs the model that the provided image represents the user’s current viewpoint, ensuring that directions are based on this perspective rather than inferred or calculated locations. This ensures that GPT-4V generates guidance that is directly relevant and accurate to the user’s perspective. Specifying the starting point in the prompt ensures the model interprets the user’s current viewpoint correctly, enabling it to provide more precise and contextually appropriate navigation instructions.

\paragraph{Integrating Direction Description}
The second part of our system prompt involves incorporating direction descriptions. This will allow GPT-4V to offer clear and actionable navigation routes. By providing detailed directions such as ``turn left at the next intersection'' or ``walk straight for 100 meters,'' the model can give precise and practical instructions that users can easily follow. Integrating direction descriptions is crucial for providing practical navigation assistance, as it transforms general guidance into specific, step-by-step instructions.

\paragraph{Filtering Irrelevant Details}
The third part of our system prompt design focuses on filtering out irrelevant details. This helps GPT-4V ignore information that is not useful for people with blindness, such as the color, shape, and visual size of objects. By removing these unnecessary details, the model can focus on providing information that is truly helpful for navigation, such as landmarks, obstacles, and directional cues. This aspect of the prompt prevents information overload and ensures that the guidance remains focused and relevant to the user's needs. By streamlining the information provided, we improve the model's ability to deliver precise and efficient navigation assistance, making it easier for users to follow.

\paragraph{Preventing Hallucination}
The final part of our system prompt design aims to prevent hallucinations. This directs GPT-4V to avoid generating information that is not based on the input image or verifiable data. Hallucination refers to the model providing inaccurate or fabricated directions or details. By explicitly instructing the model to refrain from hallucinating, we aim to ensure the reliability and trustworthiness of the navigation assistance provided. This aspect of the prompt emphasizes the importance of grounding the model's responses in visual and contextual information, thereby minimizing the risk of misleading or incorrect guidance. By addressing hallucination, we enhance the overall safety and effectiveness of GPT-4V in navigation tasks, ensuring that users receive dependable and accurate directions.

\subsection{Case Study}
\label{app:case_study}
In the navigation task, VLMs showed significant differences in their ability to interpret the same scene, as shown in Figure~\ref{fig_navigation_example15}. This example, in particular, highlights the critical ``brittleness'' and ``unreliability'' that our multi-dimensional evaluation rubric was designed to catch. At first glance, Claude's response appears detailed. It correctly identifies the ``backpack or bag'' and provides a route (``walk straight ahead... 8 steps''). However, its core instruction that the chair is ``directly in front of you'' is imprecise and potentially incorrect, as it ignores the intervening table. Under our rubric, this response was rated `Correct' on `Route' (as it provided one) and `Correct' on `Obstacles' (it did not hallucinate). However, it was rated `Incorrect' on `Destination,' as its imprecise guidance fails to get the user reliably to the specific chair. Similarly, GPT-4o's response correctly uses the table as a landmark (``Move forward until you reach the table...''). However, it also includes a critical failure: it hallucinates a ``wet floor sign'' that is not present in the image. This is not a sign of environmental safety, but rather a dangerous failure of trustworthiness that could mislead a blind person. Therefore, this response was rated 'Incorrect' on `Obstacles.'

In comparison, the performance of Gemini-1.5 Pro is inferior. Its judgment of the status of the chair in the picture is not accurate enough. It mistakenly believes that the chair is ``occupied by items'' and ``no other chairs are available'' and does not provide any substantive navigation guidance, even suggesting users provide additional images for more information. This overly conservative strategy, although it may reduce false guidance, is inefficient in navigation tasks that require immediate resolution.
Judging from the experimental results, Claude performs better in task execution capabilities. The instructions Claude generates are detailed and precise, and they can efficiently complete navigation goals. GPT-4o is also characterized by concise and clear path planning combined with detailed environmental safety reminders, which is suitable for mission-focused applications in safe scenarios. However, Gemini performs poorly in navigation tasks and cannot meet actual mission requirements. These differences suggest that model suitability needs to be chosen based on task complexity and execution requirements.
\input{figures/navigation_example15}

\end{document}

%% file: figures/counting.tex
\begin{figure*}[htb!]
  \centering
  \includegraphics[trim={0 0 0 0},clip,width=16cm]{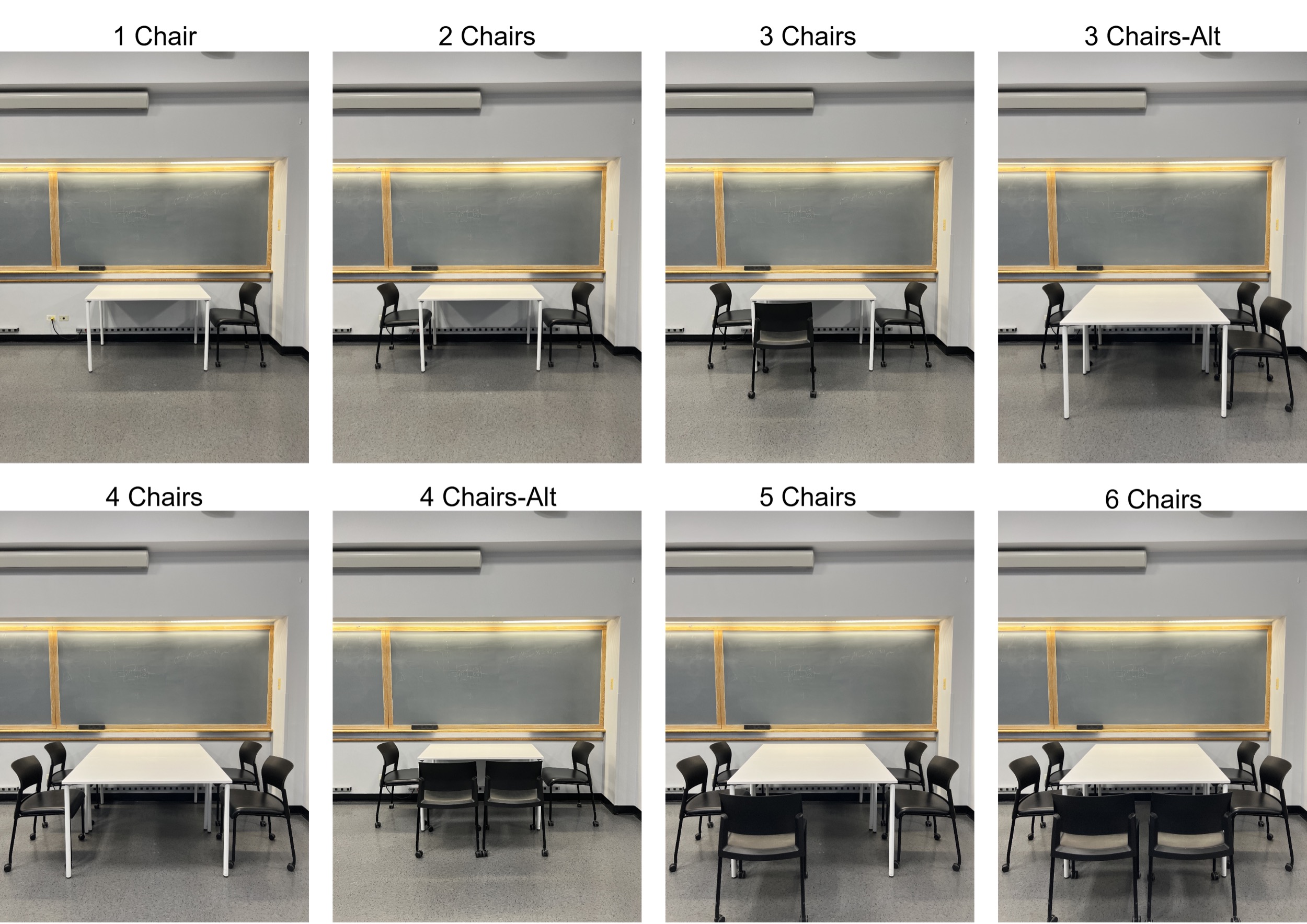}
  \caption{Evaluation examples for the fundamental counting task, images feature one to six chairs, with varying arrangements for scenarios involving three and four chairs.}
  \label{fig_counting}
\end{figure*}

%% file: tables/counting.tex
\begin{table*}[htb!]
    \centering
    \caption{Counting task results: Accuracy, mean, and variance of model predictions over 100 evaluations per test image, ranging from 1 to 6 chairs. The mean represents the average number of chairs predicted by the models, and the variance indicates the consistency of these predictions.}
    \label{tab:counting}
    \resizebox{\textwidth}{!}{
    \begin{tabular}{l ccc ccc ccc ccc ccc ccc}
        \toprule
        \multicolumn{1}{l}{Acc., Mean, Var.} & \multicolumn{3}{c}{\textbf{GPT-4V}} & \multicolumn{3}{c}{\textbf{GPT-4o}} & \multicolumn{3}{c}{\textbf{Gemini-1.5 Pro}} & \multicolumn{3}{c}{\textbf{Llava-mistral}} & \multicolumn{3}{c}{\textbf{Llava-qwen}} & \multicolumn{3}{c}{\textbf{Claude-3.5-sonnet}} \\
        \cmidrule(lr){2-4} \cmidrule(lr){5-7} \cmidrule(lr){8-10} \cmidrule(lr){11-13} \cmidrule(lr){14-16} \cmidrule(lr){17-19}
        \textbf{1 Chair} & \gradientcell{99}{0}{100}{red}{green}{66} & 1.00 & 0.01 & \gradientcell{100}{0}{100}{red}{green}{66} & \textbf{1.00} & \textbf{0.00} & \gradientcell{99}{0}{100}{red}{green}{66} & 0.99 & 0.01 & \gradientcell{38}{0}{100}{red}{green}{66} & 1.60 & 0.26 & \gradientcell{58}{0}{100}{red}{green}{66} & 1.44 & 0.31 & \gradientcell{100}{0}{100}{red}{green}{66} & \textbf{1.00} & \textbf{0.00} \\
        \textbf{2 Chairs} & \gradientcell{99}{0}{100}{red}{green}{66} & 2.01 & 0.01 & \gradientcell{100}{0}{100}{red}{green}{66} & \textbf{2.00} & \textbf{0.00} & \gradientcell{99}{0}{100}{red}{green}{66} & 1.98 & 0.04 & \gradientcell{83}{0}{100}{red}{green}{66} & 2.11 & 0.15 & \gradientcell{96}{0}{100}{red}{green}{66} & 2.07 & 0.13 & \gradientcell{100}{0}{100}{red}{green}{66} & \textbf{2.00} & \textbf{0.00} \\
        \textbf{3 Chairs} & \gradientcell{60}{0}{100}{red}{green}{66} & 3.41 & 0.26 & \gradientcell{100}{0}{100}{red}{green}{66} & \textbf{3.00} & \textbf{0.00} & \gradientcell{7}{0}{100}{red}{green}{66} & 2.02 & 0.16 & \gradientcell{56}{0}{100}{red}{green}{66} & 2.56 & 0.31 & \gradientcell{57}{0}{100}{red}{green}{66} & 2.67 & 0.32 & \gradientcell{0}{0}{100}{red}{green}{66} & 2.00 & 0.00 \\
        \textbf{3 Chairs-Alt} & \gradientcell{28}{0}{100}{red}{green}{66} & 2.28 & 0.20 & \gradientcell{77}{0}{100}{red}{green}{66} & 3.23 & 0.18 & \gradientcell{14}{0}{100}{red}{green}{66} & 1.45 & 1.36 & \gradientcell{39}{0}{100}{red}{green}{66} & 2.53 & 0.39 & \gradientcell{84}{0}{100}{red}{green}{66} & 3.00 & 0.16 & \gradientcell{100}{0}{100}{red}{green}{66} & \textbf{3.00} & \textbf{0.00} \\
        \textbf{4 Chairs} & \gradientcell{3}{0}{100}{red}{green}{66} & 3.03 & 0.03 & \gradientcell{98}{0}{100}{red}{green}{66} & 4.02 & 0.02 & \gradientcell{45}{0}{100}{red}{green}{66} & 2.22 & 4.84 & \gradientcell{70}{0}{100}{red}{green}{66} & 3.74 & 0.42 & \gradientcell{100}{0}{100}{red}{green}{66} & \textbf{4.00} & \textbf{0.00} & \gradientcell{100}{0}{100}{red}{green}{66} & \textbf{4.00} & \textbf{0.00} \\
        \textbf{4 Chairs-Alt} & \gradientcell{61}{0}{100}{red}{green}{66} & 3.67 & 0.28 & \gradientcell{99}{0}{100}{red}{green}{66} & 4.01 & 0.01 & \gradientcell{32}{0}{100}{red}{green}{66} & 1.28 & 3.52 & \gradientcell{61}{0}{100}{red}{green}{66} & 3.57 & 0.33 & \gradientcell{100}{0}{100}{red}{green}{66} & \textbf{4.00} & \textbf{0.00} & \gradientcell{100}{0}{100}{red}{green}{66} & \textbf{4.00} & \textbf{0.00} \\
        \textbf{5 Chairs} & \gradientcell{83}{0}{100}{red}{green}{66} & \textbf{4.95} & \textbf{0.17} & \gradientcell{82}{0}{100}{red}{green}{66} & 4.64 & 1.99 & \gradientcell{36}{0}{100}{red}{green}{66} & 3.30 & 6.70 & \gradientcell{5}{0}{100}{red}{green}{66} & 3.62 & 0.56 & \gradientcell{29}{0}{100}{red}{green}{66} & 5.67 & 0.26 & \gradientcell{0}{0}{100}{red}{green}{66} & 4.00 & 0.00 \\
        \textbf{6 Chairs} & \gradientcell{29}{0}{100}{red}{green}{66} & 5.20 & 0.40 & \gradientcell{92}{0}{100}{red}{green}{66} & 6.12 & 0.19 & \gradientcell{25}{0}{100}{red}{green}{66} & 1.84 & 7.35 & \gradientcell{0}{0}{100}{red}{green}{66} & 3.85 & 0.43 & \gradientcell{80}{0}{100}{red}{green}{66} & 6.36 & 0.56 & \gradientcell{100}{0}{100}{red}{green}{66} & \textbf{6.00} & \textbf{0.00} \\
        \midrule
        \textbf{Overall} & \gradientcell{58}{0}{100}{red}{green}{66} & - & - & \gradientcell{94}{0}{100}{red}{green}{66} & - & - & \gradientcell{45}{0}{100}{red}{green}{66} & - & - & \gradientcell{44}{0}{100}{red}{green}{66} & - & - & \gradientcell{76}{0}{100}{red}{green}{66} & - & - & \gradientcell{75}{0}{100}{red}{green}{66} & - & - \\
        \bottomrule 
    \end{tabular}
    }

\end{table*}

%% file: figures/spatial_reasoning_relative.tex
\begin{figure*}[h!]
  \centering
  \includegraphics[trim={0 0 0 0},clip,width=16cm]{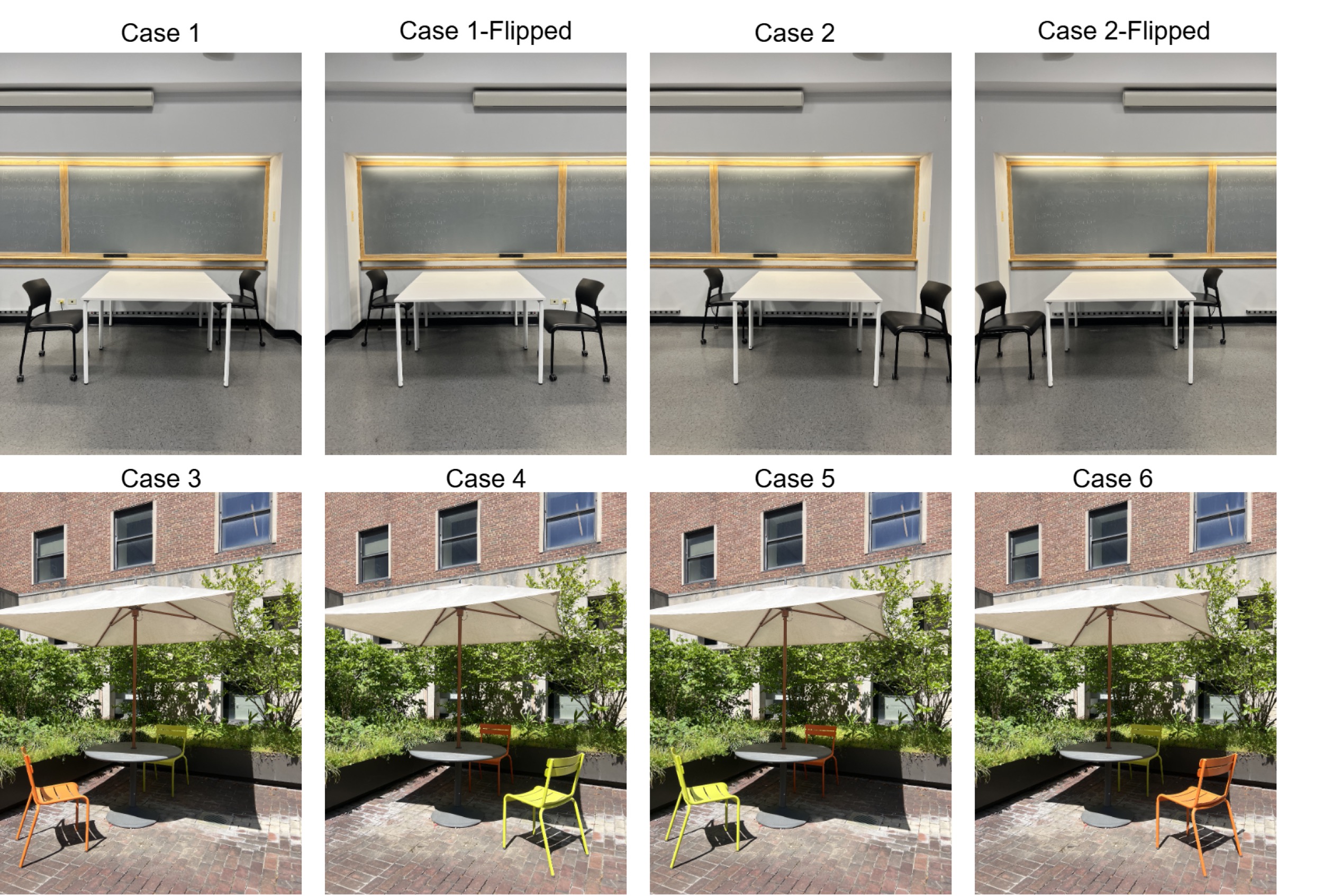}
  \caption{Evaluation examples for the fundamental relative spatial reasoning task: Images feature chairs at varying distances from the viewpoint, with both different and identical chair styles. Cases 1 and 2 are also flipped to investigate potential biases in model predictions.}
  \label{fig_relative_distance}
\end{figure*}

%% file: tables/comparative.tex
\begin{table*}[h!]
    \centering
    \setlength{\tabcolsep}{4pt} %
    \caption{Relative spatial reasoning task results: Accuracy percentages from 100 evaluations for each test case. The table highlights model performance across different scenarios from low accuracy (red) to high accuracy (green).}
    \label{tab:comparative}
    \resizebox{\textwidth}{!}{ %
    \begin{tabular}{l cccccc}
        \toprule
        \multicolumn{1}{l}{Accuracy (\%)} & \textbf{GPT-4V} & \textbf{GPT-4o} & \textbf{Gemini-1.5 Pro} & \textbf{Llava-mistral} & \textbf{Llava-qwen} & \textbf{Claude-3.5-sonnet} \\
        \midrule
        Case 1 & \gradientcell{97}{0}{100}{red}{green}{66} & \gradientcell{100}{0}{100}{red}{green}{66} & \gradientcell{36}{0}{100}{red}{green}{66} & \gradientcell{34}{0}{100}{red}{green}{66} & \gradientcell{61}{0}{100}{red}{green}{66} & \gradientcell{77}{0}{100}{red}{green}{66} \\
        Case 1-Flipped & \gradientcell{61}{0}{100}{red}{green}{66} & \gradientcell{100}{0}{100}{red}{green}{66} & \gradientcell{77}{0}{100}{red}{green}{66} & \gradientcell{72}{0}{100}{red}{green}{66} & \gradientcell{50}{0}{100}{red}{green}{66} & \gradientcell{12}{0}{100}{red}{green}{66} \\
        Case 2 & \gradientcell{13}{0}{100}{red}{green}{66} & \gradientcell{100}{0}{100}{red}{green}{66} & \gradientcell{87}{0}{100}{red}{green}{66} & \gradientcell{53}{0}{100}{red}{green}{66} & \gradientcell{60}{0}{100}{red}{green}{66} & \gradientcell{12}{0}{100}{red}{green}{66} \\
        Case 2-Flipped & \gradientcell{81}{0}{100}{red}{green}{66} & \gradientcell{99}{0}{100}{red}{green}{66} & \gradientcell{47}{0}{100}{red}{green}{66} & \gradientcell{94}{0}{100}{red}{green}{66} & \gradientcell{74}{0}{100}{red}{green}{66} & \gradientcell{30}{0}{100}{red}{green}{66} \\
        Case 3 & \gradientcell{100}{0}{100}{red}{green}{66} & \gradientcell{100}{0}{100}{red}{green}{66} & \gradientcell{100}{0}{100}{red}{green}{66} & \gradientcell{37}{0}{100}{red}{green}{66} & \gradientcell{98}{0}{100}{red}{green}{66} & \gradientcell{99}{0}{100}{red}{green}{66} \\
        Case 4 & \gradientcell{98}{0}{100}{red}{green}{66} & \gradientcell{100}{0}{100}{red}{green}{66} & \gradientcell{28}{0}{100}{red}{green}{66} & \gradientcell{81}{0}{100}{red}{green}{66} & \gradientcell{94}{0}{100}{red}{green}{66} & \gradientcell{100}{0}{100}{red}{green}{66} \\
        Case 5 & \gradientcell{74}{0}{100}{red}{green}{66} & \gradientcell{99}{0}{100}{red}{green}{66} & \gradientcell{100}{0}{100}{red}{green}{66} & \gradientcell{62}{0}{100}{red}{green}{66} & \gradientcell{96}{0}{100}{red}{green}{66} & \gradientcell{24}{0}{100}{red}{green}{66} \\
        Case 6 & \gradientcell{100}{0}{100}{red}{green}{66} & \gradientcell{100}{0}{100}{red}{green}{66} & \gradientcell{60}{0}{100}{red}{green}{66} & \gradientcell{66}{0}{100}{red}{green}{66} & \gradientcell{94}{0}{100}{red}{green}{66} & \gradientcell{100}{0}{100}{red}{green}{66} \\
        \midrule
        Overall & \gradientcell{78}{0}{100}{red}{green}{66} & \gradientcell{100}{0}{100}{red}{green}{66} & \gradientcell{67}{0}{100}{red}{green}{66} & \gradientcell{62}{0}{100}{red}{green}{66} & \gradientcell{78}{0}{100}{red}{green}{66} & \gradientcell{57}{0}{100}{red}{green}{66}\\
        \bottomrule
    \end{tabular}
    }
\end{table*}

%% file: figures/spatial_reasoning_analysis.tex
\begin{figure*}[htb!]
  \centering
  \includegraphics[trim={0 0 0 0},clip,width=16cm]{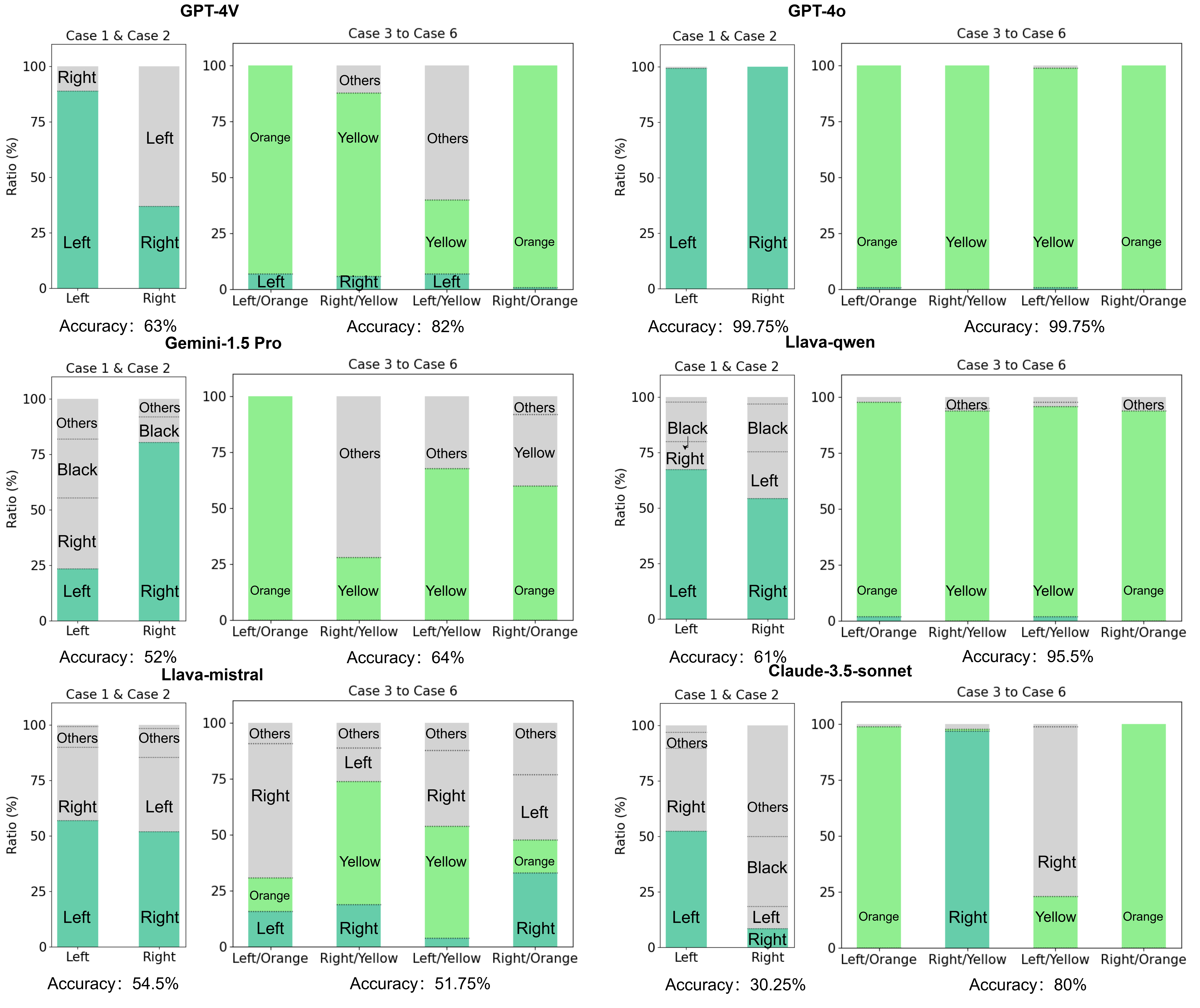}
  \caption{Spatial reasoning keyword analysis: The x-axis represents the correct answer, while the bars indicate the ratio of responses extracted from model outputs. Dark green bars denote optimal answers for pBLV applications, using spatial terms (e.g., ``left'', ``right'') to describe positional relationships. Light green bars represent suboptimal answers that rely on color-based descriptions (e.g., ``orange'', ``yellow'') but demonstrate correct spatial reasoning.}
  \label{fig_spatial_analysis}
\end{figure*}

%% file: figures/commonsense_reasoning.tex
\begin{figure*}[htb!]
  \centering
  \includegraphics[trim={0 0 0 0},clip,width=16cm]{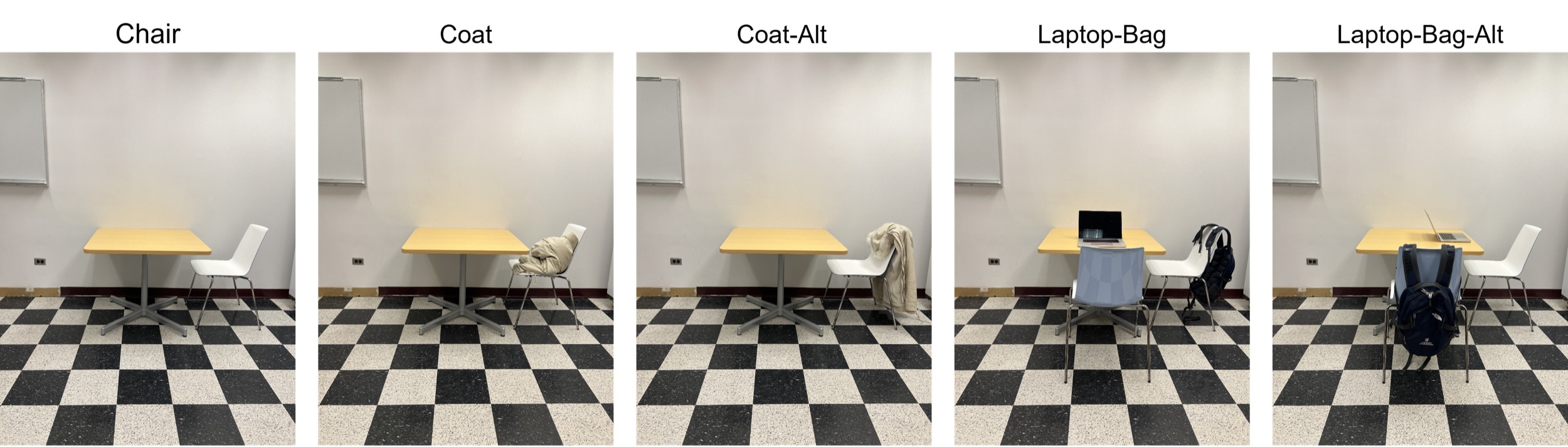}
  \caption{Evaluation examples for the fundamental commonsense reasoning task: Images show chairs that are either occupied or available, with various objects such as coats hanging on the chair or laptops placed on surfaces to indicate occupancy.}
  \label{fig_commonsense_reasoning}
\end{figure*}

%% file: tables/commonmense.tex
\begin{table*}[htb!]
    \centering
    \caption{Common-sense task results: Accuracy percentages from 100 evaluations for each test case. The table highlights model performance across different scenarios from low accuracy (red) to high accuracy (green).}
    \label{tab:commonsense}
    \resizebox{\textwidth}{!}{ %
    \begin{tabular}{l cccccc} 
        \toprule
        \multicolumn{1}{l}{Accuracy (\%)} & \textbf{GPT-4V} 
        & \textbf{GPT-4o} 
        & \textbf{Gemini-1.5 Pro} 
        & \textbf{Llava-mistral} 
        & \textbf{Llava-qwen}
        & \textbf{Claude-3.5-sonnet} \\ %
        \midrule
        Chair & \gradientcell{100}{0}{100}{red}{green}{66} & \gradientcell{100}{0}{100}{red}{green}{66} & \gradientcell{72}{0}{100}{red}{green}{66} & \gradientcell{100}{0}{100}{red}{green}{66} & \gradientcell{100}{0}{100}{red}{green}{66} & \gradientcell{100}{0}{100}{red}{green}{66} \\
        Coat & \gradientcell{49}{0}{100}{red}{green}{66} & \gradientcell{100}{0}{100}{red}{green}{66} & \gradientcell{100}{0}{100}{red}{green}{66} & \gradientcell{1}{0}{100}{red}{green}{66} & \gradientcell{2}{0}{100}{red}{green}{66} & \gradientcell{0}{0}{100}{red}{green}{66} \\
        Coat-Alt & \gradientcell{29}{0}{100}{red}{green}{66} & \gradientcell{100}{0}{100}{red}{green}{66} & \gradientcell{100}{0}{100}{red}{green}{66} & \gradientcell{3}{0}{100}{red}{green}{66} & \gradientcell{0}{0}{100}{red}{green}{66} & \gradientcell{0}{0}{100}{red}{green}{66} \\
        Laptop-Backpack & \gradientcell{6}{0}{100}{red}{green}{66} & \gradientcell{95}{0}{100}{red}{green}{66} & \gradientcell{42}{0}{100}{red}{green}{66} & \gradientcell{4}{0}{100}{red}{green}{66} & \gradientcell{1}{0}{100}{red}{green}{66} & \gradientcell{0}{0}{100}{red}{green}{66} \\
        Laptop-Backpack-Alt & \gradientcell{18}{0}{100}{red}{green}{66} & \gradientcell{7}{0}{100}{red}{green}{66} & \gradientcell{0}{0}{100}{red}{green}{66} & \gradientcell{2}{0}{100}{red}{green}{66} & \gradientcell{6}{0}{100}{red}{green}{66} & \gradientcell{0}{0}{100}{red}{green}{66} \\
        \midrule
        Overall & \gradientcell{40}{0}{100}{red}{green}{66} & \gradientcell{80}{0}{100}{red}{green}{66} & \gradientcell{63}{0}{100}{red}{green}{66} & \gradientcell{22}{0}{100}{red}{green}{66} & \gradientcell{22}{0}{100}{red}{green}{66} & \gradientcell{20}{0}{100}{red}{green}{66} \\
        \bottomrule
    \end{tabular}
    }
\end{table*}

%% file: figures/navigation_all.tex
\begin{figure*}[htb!]
  \centering
  \includegraphics[trim={0 0 0 0},clip,width=16cm]{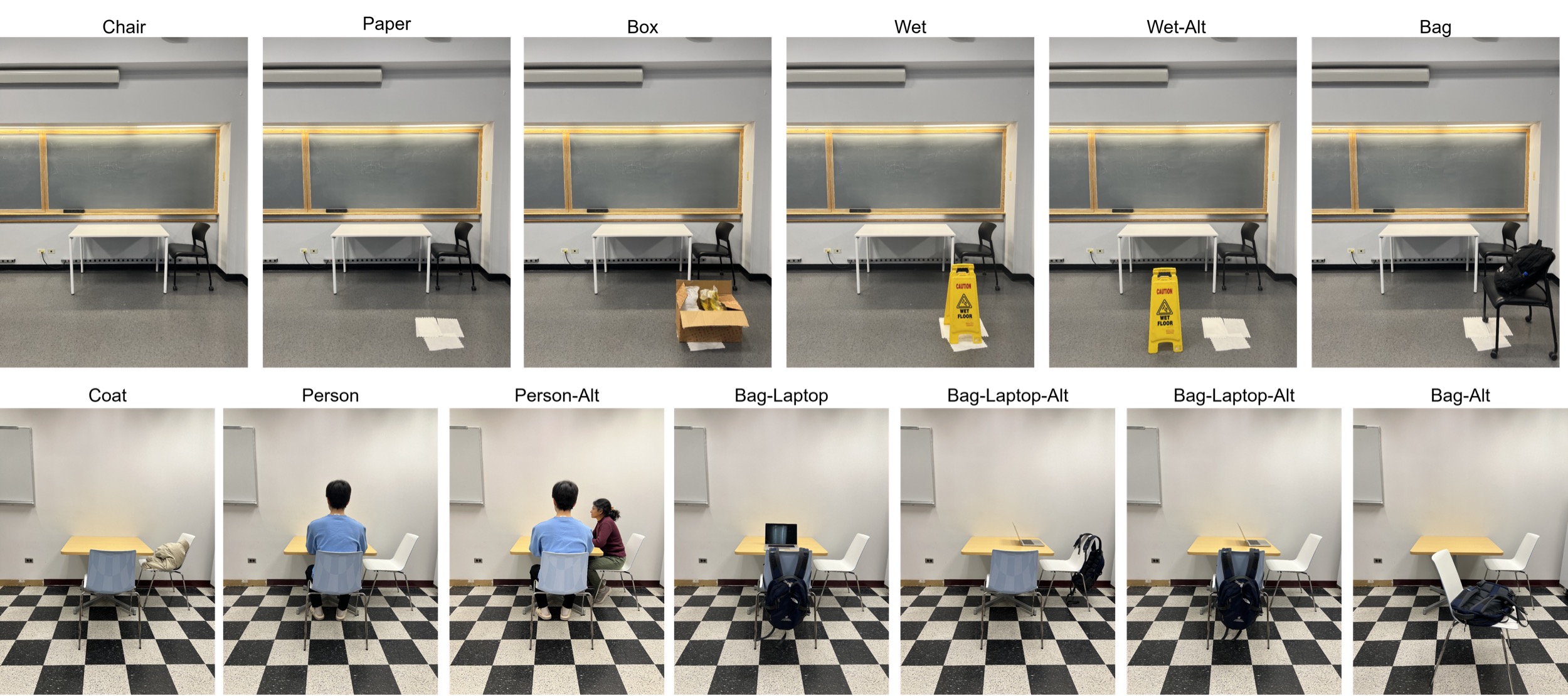}
  \caption{Evaluation examples for navigation tasks: Scenarios include obstacles placed along the path to the chair or objects placed on the chair and corresponding tabletop. Obstacle and object types, such as backpacks, coats, and boxes, simulate complex real-life scenarios to assess the model’s ability to identify empty chairs, plan reasonable navigation paths, and avoid obstacles effectively.}
  \label{fig_navigation}
\end{figure*}

%% file: tables/navigation.tex
\begin{table}[htb!]
    \centering
    \caption{Navigation task results: Accuracy of navigation tasks, including destination identification, route planning, and obstacle detection, as judged by assessment annotators.}
    \label{tab:navigation}
    \resizebox{\columnwidth}{!}{ %
    \begin{tabular}{l ccc}
        \toprule
        \multicolumn{1}{l}{Accuracy (\%)} & \textbf{Destination} & \textbf{Route} & \textbf{Obstacles} \\
        \midrule
        GPT-4o & \gradientcell{81}{0}{100}{red}{green}{66} & \gradientcell{92}{0}{100}{red}{green}{66} & \gradientcell{84}{0}{100}{red}{green}{66} \\
        Gemini-1.5 Pro & \gradientcell{30}{0}{100}{red}{green}{66} & \gradientcell{13}{0}{100}{red}{green}{66} & \gradientcell{64}{0}{100}{red}{green}{66} \\
        Llava-mistral & \gradientcell{66}{0}{100}{red}{green}{66} & \gradientcell{67}{0}{100}{red}{green}{66} & \gradientcell{74}{0}{100}{red}{green}{66} \\
        Llava-qwen & \gradientcell{40}{0}{100}{red}{green}{66} & \gradientcell{38}{0}{100}{red}{green}{66} & \gradientcell{62}{0}{100}{red}{green}{66} \\
        Claude-3.5-sonnet & \gradientcell{85}{0}{100}{red}{green}{66} & \gradientcell{87}{0}{100}{red}{green}{66} & \gradientcell{82}{0}{100}{red}{green}{66} \\
        \bottomrule
    \end{tabular}
    }
\end{table}

%% file: tables/discussion.tex
\begin{table}[htb!]
    \centering
    \caption{Overall accuracy performance summary of models across the three fundamental tasks: counting, relative spatial reasoning, and commonsense reasoning.}
    \label{tab:discussion}
    \resizebox{\columnwidth}{!}{ %
    \begin{tabular}{l ccc}
        \toprule
        \multicolumn{1}{l}{Overall Accuracy(\%)} & \textbf{Counting} & \textbf{Spatial} & \textbf{Commonsense} \\
        \midrule
        GPT-4V & \gradientcell{58}{0}{100}{red}{green}{66} & \gradientcell{78}{0}{100}{red}{green}{66} & \gradientcell{40}{0}{100}{red}{green}{66} \\
        GPT-4o & \gradientcell{94}{0}{100}{red}{green}{66} & \gradientcell{100}{0}{100}{red}{green}{66} & \gradientcell{80}{0}{100}{red}{green}{66} \\
        Gemini-1.5 Pro & \gradientcell{45}{0}{100}{red}{green}{66} & \gradientcell{67}{0}{100}{red}{green}{66} & \gradientcell{63}{0}{100}{red}{green}{66} \\
        Llava-mistral & \gradientcell{44}{0}{100}{red}{green}{66} & \gradientcell{62}{0}{100}{red}{green}{66} & \gradientcell{22}{0}{100}{red}{green}{66} \\
        Llava-qwen & \gradientcell{76}{0}{100}{red}{green}{66} & \gradientcell{78}{0}{100}{red}{green}{66} & \gradientcell{22}{0}{100}{red}{green}{66} \\
        Claude-3.5-sonnet & \gradientcell{75}{0}{100}{red}{green}{66} & \gradientcell{57}{0}{100}{red}{green}{66} & \gradientcell{20}{0}{100}{red}{green}{66} \\
        \bottomrule
    \end{tabular}
    }
\end{table}

%% file: tables/resolution-1.tex
\begin{table*}[htb!]
    \centering
    \caption{Resolution Comparison in Counting Task}
    \label{tab:resolution-1}
    \resizebox{\textwidth}{!}{
    \begin{tabular}{l ccc ccc ccc ccc ccc ccc}
        \toprule
        \multicolumn{1}{l}{Acc. (\%), Mean, Var.} 
        & \multicolumn{6}{c}{\textbf{Llava-mistral}} 
        & \multicolumn{6}{c}{\textbf{Llava-qwen}} 
        & \multicolumn{6}{c}{\textbf{Claude-3.5-sonnet}} \\
        \cmidrule(lr){2-7} \cmidrule(lr){8-13} \cmidrule(lr){14-19}
        \multicolumn{1}{l}{Resolution} 
        & \multicolumn{3}{c}{Original}
        & \multicolumn{3}{c}{252x336}
        & \multicolumn{3}{c}{Original}
        & \multicolumn{3}{c}{288x384} 
        & \multicolumn{3}{c}{Original}
        & \multicolumn{3}{c}{150x200}\\
        \cmidrule(lr){2-4} \cmidrule(lr){5-7} \cmidrule(lr){8-10} \cmidrule(lr){11-13} \cmidrule(lr){14-16} \cmidrule(lr){17-19}
        1 Chair & 38 & 1.60 & 0.26 & 31 & 1.69 & 0.22 & 40 & 1.72 & 0.67 & 58 & 1.44 & 0.31 & 100 & 1.00 & 0.00 & 99 & 0.99 & 0.01 \\
        2 Chairs & 83 & 2.11 & 0.15 & 99 & 2.01 & 0.01 & 99 & 2.02 & 0.04 & 96 & 2.07 & 0.13 & 100 & 2.00 & 0.00 & 100 & 2.00 & 0.00 \\
        3 Chairs & 56 & 2.56 & 0.31 & 23 & 2.23 & 0.18 & 3 & 2.07 & 0.11 & 57 & 2.67 & 0.32 & 0 & 2.00 & 0.00 & 0 & 2.00 & 0.00 \\
        3 Chairs-Alt & 39 & 2.53 & 0.39 & 65 & 3.09 & 0.35 & 19 & 2.19 & 0.16 & 84 & 3.00 & 0.16 & 100 & 3.00 & 0.00 & 0 & 2.00 & 0.00 \\
        4 Chairs & 70 & 3.74 & 0.42 & 65 & 3.66 & 0.27 & 30 & 2.91 & 0.75 & 100 & 4.00 & 0.00 & 100 & 4.00 & 0.00 & 100 & 4.00 & 0.00 \\
        4 Chairs-Alt & 61 & 3.57 & 0.33 & 37 & 3.32 & 0.32 & 8 & 2.22 & 0.33 & 100 & 4.00 & 0.00 & 100 & 4.00 & 0.00 & 100 & 4.00 & 0.00 \\
        5 Chairs & 5 & 3.62 & 0.56 & 10 & 4.11 & 0.20 & 3 & 3.54 & 0.61 & 29 & 5.67 & 0.26 & 0 & 4.00 & 0.00 & 0 & 4.06 & 0.12 \\
        6 Chairs & 0 & 3.85 & 0.43 & 6 & 4.36 & 0.43 & 7 & 4.13 & 2.11 & 80 & 6.36 & 0.56 & 100 & 6.00 & 0.00 & 10 & 4.20 & 0.36 \\
        \midrule
        \textbf{Overall} 
        & 44 & - & -
        & 42 & - & -
        & 26 & - & -
        & 78 & - & -
        & 86 & - & -
        & 58 & - & - \\
        \bottomrule
    \end{tabular}
    }
\end{table*}

%% file: tables/resolution-2.tex
\begin{table*}[htb!]
    \centering
    \caption{Resolution Comparison in Spatial Reasoning Task}
    \label{tab:resolution-2}
    \begin{tabular}{l cccccc}
        \toprule
        \multicolumn{1}{l}{Accuracy (\%)} 
        & \multicolumn{2}{c}{\textbf{Llava-mistral}} 
        & \multicolumn{2}{c}{\textbf{Llava-qwen}} 
        & \multicolumn{2}{c}{\textbf{Claude-3.5-sonnet}} \\
        \cmidrule(lr){2-3} \cmidrule(lr){4-5} \cmidrule(lr){6-7}
        \multicolumn{1}{l}{Resolution} 
        & \footnotesize{Original} 
        & \footnotesize{252x336} 
        & \footnotesize{Original} 
        & \footnotesize{288x384} 
        & \footnotesize{Original} 
        & \footnotesize{150x200} \\
        \midrule
        Case 1 & 34 & 80 & 11 & 61 & 77 & 41 \\
        Case 1-Flipped & 72 & 18 & 41 & 74 & 29 & 76 \\
        Case 2 & 53 & 47 & 85 & 50 & 12 & 61 \\
        Case 2-Flipped & 94 & 69 & 38 & 60 & 12 & 67 \\
        Case 3 & 37 & 99 & 93 & 98 & 99 & 3 \\
        Case 4 & 62 & 95 & 87 & 94 & 100 & 5 \\
        Case 5 & 66 & 96 & 55 & 96 & 24 & 16 \\
        Case 6 & 81 & 87 & 99 & 94 & 100 & 58 \\
        \midrule
        \textbf{Overall} 
        & 62 & 74 
        & 64 & 78 
        & 57 & 41 \\
        \bottomrule
    \end{tabular}
\end{table*}

%% file: tables/resolution-3.tex
\begin{table*}[htb!]
    \centering
    \caption{Resolution Comparison in Commonsense Reasoning Task}
    \label{tab:resolution-3}
    \begin{tabular}{l cccccc}
        \toprule
        \multicolumn{1}{l}{Accuracy (\%)} 
        & \multicolumn{2}{c}{\textbf{Llava-mistral}} 
        & \multicolumn{2}{c}{\textbf{Llava-qwen}} 
        & \multicolumn{2}{c}{\textbf{Claude-3.5-sonnet}} \\
        \cmidrule(lr){2-3} \cmidrule(lr){4-5} \cmidrule(lr){6-7}
        \multicolumn{1}{l}{Resolution} 
        & \footnotesize{Original} 
        & \footnotesize{252x336} 
        & \footnotesize{Original} 
        & \footnotesize{288x384} 
        & \footnotesize{Original} 
        & \footnotesize{150x200} \\
        \midrule
        Chair & 100 & 99 & 99 & 100 & 100 & 100 \\
        Coat & 1 & 0 & 6 & 2 & 0 & 0 \\
        Coat-Alt & 3 & 0 & 2 & 0 & 0 & 100 \\
        Laptop-Backpack & 4 & 0 & 3 & 1 & 0 & 0 \\
        Laptop-Backpack-Alt & 2 & 0 & 10 & 6 & 0 & 0 \\
        \midrule
        \textbf{Overall} 
        & 22 & 20 
        & 24 & 22 
        & 20 & 40 \\
        \bottomrule
    \end{tabular}
\end{table*}

%% file: figures/system_prompt.tex
\begin{figure*}[htb!]
  \centering
  \includegraphics[trim={0 0 0 0},clip,width=16cm]{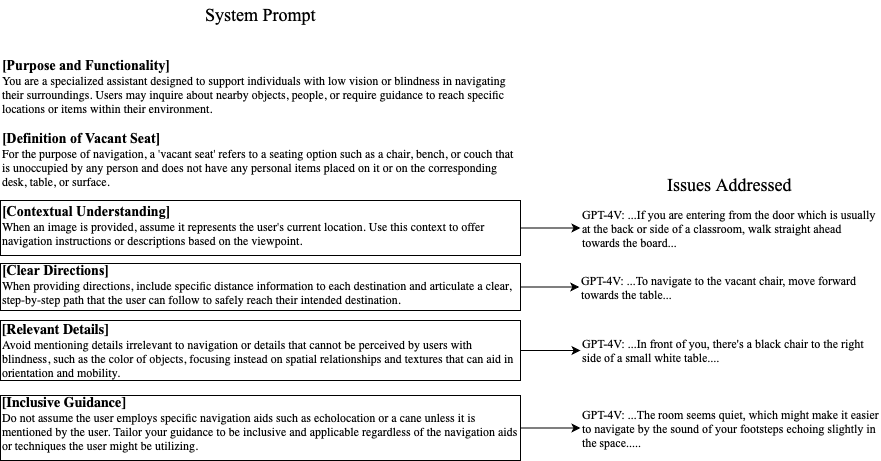}
  \caption{ The system prompt design is presented, which optimizes the model performance in navigation tasks through six parts: clear purpose and function, definition of vacant seats, context understanding, clear guidance, filtering of relevant details, and inclusive guidance. The prompt design is intended to help low-vision or visually impaired users provide accurate and effective navigation support, such as identifying the target location, planning a clear path, and providing practical descriptions of the environment, while avoiding redundant information or assumptions that do not meet user needs, thereby improving the practicality and reliability of the model in actual navigation scenarios.}
  \label{fig_system_prompt}
\end{figure*}

%% file: figures/navigation_example15.tex
\begin{figure}[H]
  \centering
  \includegraphics[trim={0 0 0 0},clip,width=8cm]{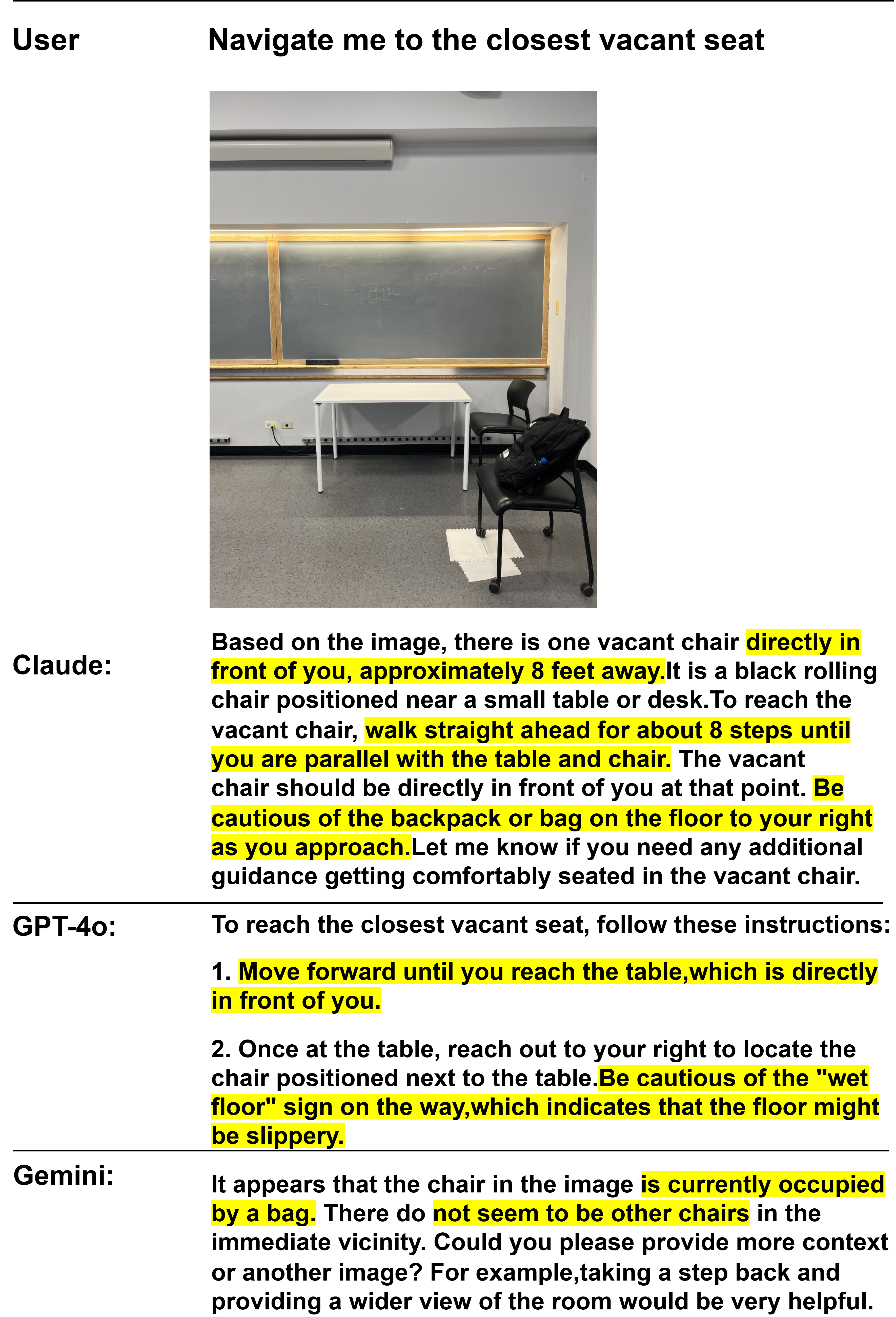}
  \caption{
  The output comparison of different VLMs in navigation tasks is shown, focusing on evaluating the differences in their ability to interpret the same image and generate navigation instructions. The figure compares the performance of each model in target positioning, path planning, and environmental information description. Some models generate precise and detailed instructions, such as clearly indicating the location of the target and potential obstacles, while providing clear path planning; while other models show obvious limitations, such as misjudging the target state, missing obstacle information, or failing to provide useful navigation guidance. Through this comparison, we can clearly see the advantages and disadvantages of each model in the navigation task, revealing its gaps in task execution capabilities and environmental understanding, and providing an important reference for further improving the model.}
  \label{fig_navigation_example15}
\end{figure}